\documentclass[letterpaper, 10 pt, conference]{ieeeconf}  

\IEEEoverridecommandlockouts                              

\usepackage{booktabs}
\usepackage{graphicx}
\usepackage{float}
\usepackage{tabularx}
\usepackage{amsfonts}
\usepackage{multirow}

\usepackage{color}
\usepackage{soul}

\title{\LARGE \bf
What Looks Good with my Sofa:\\ Multimodal Search Engine for Interior Design
}

\author{Ivona Tautkute$^{1,\, 3}$, Aleksandra Mo\.zejko$^{3}$, Wojciech Stokowiec$^{1,\, 3}$,\\ Tomasz Trzci\'{n}ski$^{2, \, 3}$, \L{}ukasz Brocki$^{1}$ and Krzysztof Marasek$^{1}$
\thanks{$^{1}$Polish-Japanese Academy of Information Technology, Warsaw, Poland}
\thanks{$^{2}$Warsaw University of Technology, Warsaw, Poland}
\thanks{$^{3}$Tooploox, Warsaw, Poland.}%
}

\begin{document}
\maketitle
\thispagestyle{empty}
\pagestyle{empty}

\begin{abstract}
In this paper, we propose a multi-modal search engine for interior design that combines visual and textual queries. The goal of our engine is to retrieve interior objects, {\it e.g.} furniture or wall clocks, that share visual and aesthetic similarities with the query. Our search engine allows the user to take a photo of a room and retrieve with a high recall a list of items identical or visually similar to those present in the photo. Additionally, it allows to return other items that aesthetically and stylistically fit well together. To achieve this goal, our system blends the results obtained using textual and visual modalities. Thanks to this blending strategy, we increase the average style similarity score of the retrieved items by 11\%. Our work is implemented as a Web-based application and it is planned to be opened to the public.
\end{abstract}

\section{Introduction}

Recent advancements in the development of efficient and effective deep learning methods that rely on multi-layer neural networks have lead to impressive results obtained for many computer vision applications, such as object detection or object classification~\cite{krizh,russ}. Nevertheless, a set of challenges regarding image understanding is still to be solved, for instance training a model which is able not only to detect an object, {\it e.g.} sofa or chair, in the picture, but based on this detection suggest a table or wallpaper to match their style. This is exactly the topic of this work and the applications of such system are numerous, including but not limited to interior design augmented reality applications or e-commerce recommendation engines.


Although several methods for finding visually similar objects exist~\cite{nister,philbin}, they rather focus on the similarities related to the appearance of the objects, not their style or context. On the other hand, recently proposed textual representation called {\it word2vec}~\cite{word2vec} that is used in many text-based search engines is trained mainly using contextual information present in the training corpus. This approach allows to map words describing objects that often appear together, {\it e.g.} {\it chair} and {\it table}, to spaces where their representations are closer to each other than, {\it e.g.} {\it table} and {\it bathtub}. Therefore, one can imagine using word2vec representation for finding interior design items that correspond to the same style, as they would often appear together.  Nevertheless, textual search often falls short when applied to interior design applications, as the variety of stylistic and aesthetic descriptions, such as {\it Scandinavian style} or {\it minimalistic design}, is only known by a limited number of professional interior designers, and remains cryptic for target users of those applications.

      

\begin{figure*}[t]
\begin{center}
  \includegraphics[width=0.9\textwidth]{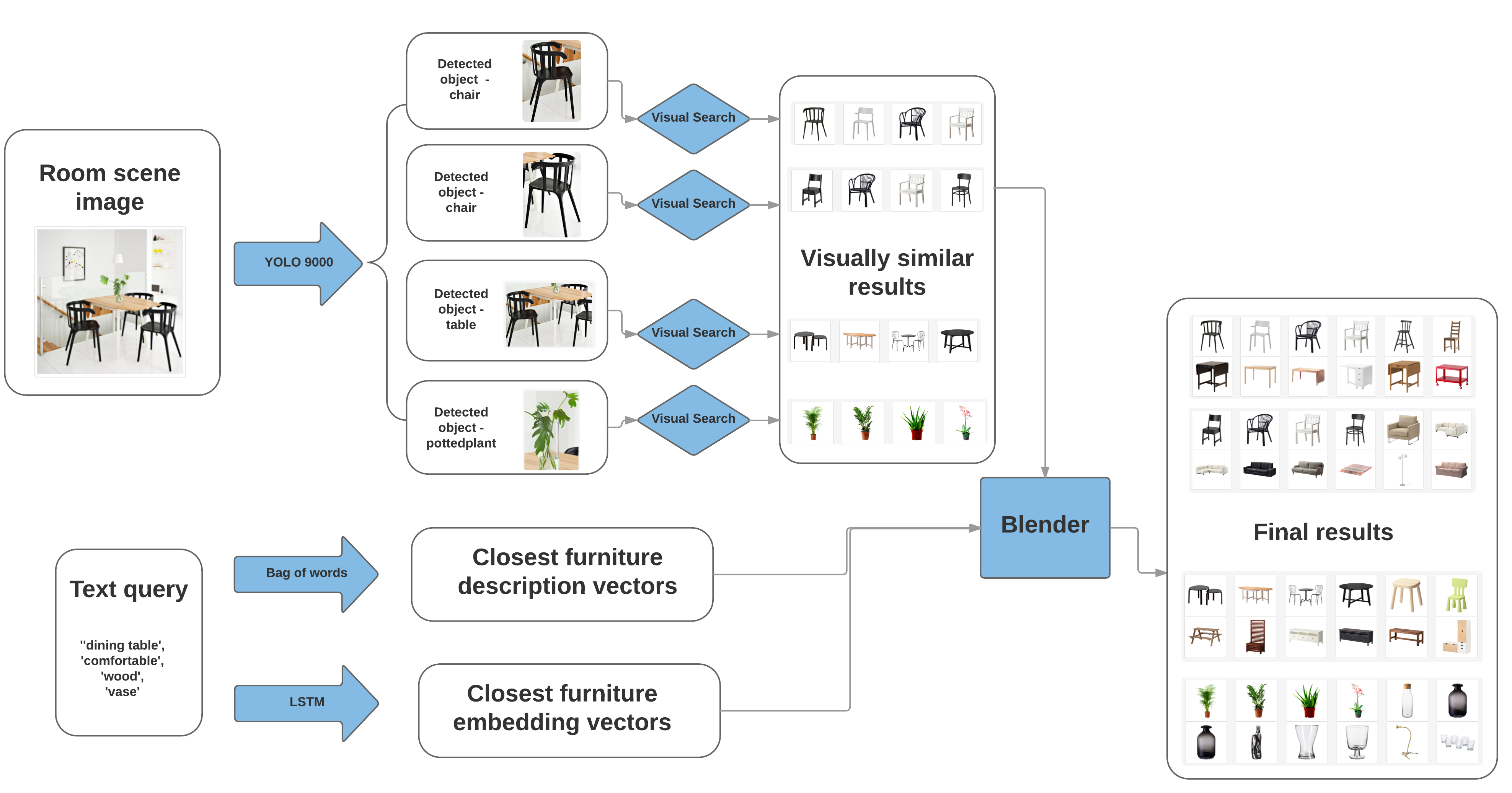} 
    \caption{High-level overview of our proposed Style Search Engine. The visual search block of our engine uses state-of-the-art object detection algorithm YOLO 9000~\cite{yolo} and the outputs of deep neural network. The textual block allows to further specify search criteria with text and increases the contextual importance of the retrieved results. Finally, by blending the visual and textual search results using similarity score in their respective feature spaces, our method significantly improves the stylistic and aesthetic similarity of the retrieved items.}
    \label{pipe}
    \end{center}
\end{figure*}

In this paper, we address the above mentioned shortcomings of visual or textual search when applied to interior design by combining the best of both worlds. More precisely, we propose a multi-modal approach to interior design search, dubbed Style Search Engine, which retrieves a list of visually similar objects enhanced with textual input from the user. Fig.~\ref{pipe} shows a high-level overview of our proposed Style Search Engine. The first building block of our engine combines state-of-the-art object detection algorithm YOLO 9000~\cite{yolo} with visual search engine based on the outputs of deep neural network. The second block allows to further specify search criteria with text and it uses this textual input for context-aware retrieval of stylistically similar objects. At final stage, our method blends the visual and textual search results using similarity score in their respective feature spaces. This leads to 11\% performance improvement in terms of style similarity of the retrieved objects.


To summarize, the contributions of this work are threefold:
\begin{itemize}

\item Firstly, we propose a multi-modal search framework that combines object detection, visual search and textual query to return a set of results that are visually and stylistically similar.

\item Secondly, we propose a new blending method for search models (image and text) that increases the quality of the results.

\item Thirdly, we implement our Style Search Engine as a working Web application with the aim of opening it to the public.
\end{itemize}

The remainder of this paper is organized in the following manner. We begin with a brief overview of the related work and then describe our Style Search Engine along with their building blocks. In Sec.~\ref{sec:dataset}, we introduce the datasets that is then used in Sec.~\ref{sec:experiments} for experiments and validation of our method. We present our Web-based application of Style Search Engine in Sec.~\ref{sec:application} and in Sec.~\ref{sec:conclusions} we conclude the paper.

\section{Related Work}

In this section, we first give an overview of the visual search methods proposed in the literature. We then discuss several approaches used in the context of textual search. Finally, we present works related to defining similarity in the context of aesthetics and style, as it directly pertains to the results obtained using our proposed method.

\subsection{Visual search}

Traditionally, image-based search methods drew their inspiration from textual retrieval systems~\cite{nister}. By using $k$-means clustering method in the space of local feature descriptors, such as SIFT~\cite{sift}, they are able to mimic textual word entities with the so-called {\it visual words}. Once the mapping from image salient keypoints to visually representative {\it words} was established, typical textual retrieval methods, such as Bag-of-Words~\cite{bow} could be used. 
Video Google~\cite{vgoogle} was one of the first visual search engines that relied on this concept. Several extensions of this concept were proposed, {\it e.g.} spatial verification~\cite{philbin} that checks for geometrical correctness of initial query and eliminates the results that are not geometrically plausible. Other descriptor pooling methods were also proposed, {\it e.g.} Fisher Vectors \cite{fisher} or VLAD \cite{vlad}. 

Successful applications of deep learning techniques in other computer vision applications have motivated researchers to apply those methods also to visual search. Although 
preliminary results did not seem promising due to lack of robustness to cropping, scaling and image clutter~\cite{gordo}, later works proved potential of those methods in the domain of image-based retrieval. For instance, by incorporation of R-MAC technique~\cite{tolias3} image representation based on the outputs of convolutional neural networks could be computed in a fixed layout of spatial regions. Many other deep architectures were  also proposed, such as siamese networks, and proved successful when applied to content-based image retrieval~\cite{prod_des}. 

Nevertheless, all of the above mentioned methods suffer from an important drawback, namely they do not take into account the contextual and stylistic similarity of the retrieved objects, which yields their application to the problem of interior design items retrieval infeasible. 

\subsection{Textual Search}

First methods proposed to address textual information retrieval have been based on token counts, {\it e.g.} \textit{Bag-of-Words} \cite{bow} or \textit{TF-IDF} \cite{tfidf}. Despite being conceptually simple and adequate to small-scale search problems, the scalability of those methods is very limited. This is due to the fact that the representation size grows with the indexed \textit{corpus} size and, in turn, causes problems with less frequent tokens. Additionally, when using such representations long sequences (documents) tend to have similar token distributions which results in lower discriminative power of the representation and lower retrieval precision. One way to avoid those problems is to apply a SVD decomposition of the token co-occurrence matrix and, hence, reduce the dimensionality of a representation vector~\cite{lsi1,lsi2}. This, however, does not address another problem commonly occurring in token-based representations, namely the fact that they are insensitive to any sequence (token) permutation. Moreover, it is not straightforward to obtain a good representation of single tokens using above mentioned methods.

To handle those shortcomings, a new type of representation called \textit{word2vec} has been proposed by Mikolov {\it et. al}~\cite{word2vec}. The proposed instances of {\it word2vec}, namely continuous Bag of Words (CBOW) and Skip-Grams, allow the token representation to be learned based on its local context. To grasp also the global context of the token, later extension of {\it word2vec} called GLoVe~\cite{glove} has been introduced. GLoVe takes advantage of information both from local context and the global co-occurrence matrix, therefore providing a powerful and discriminative representation of textual data. 

\subsection{Stylistic Similarity}
\label{subsec:stylistic_similarity}

Comparing the style similarity of two objects or scenes is one of the challenges that has to be answered when training a machine learning model for interior design retrieval application. This problem is far from being solved mainly due to the lack of a clear metric defining how to measure style similarity. Various approaches have been proposed for defining style similarity metric. Some of them focus on evaluating similarity between shapes based on their structures~\cite{kara,kaik} and measuring the differences between scales and orientations of bounding boxes. Other approach propose structure-transcending style similarity measure that accounts for element similarity, element saliency and prevalence~\cite{lun}. In this work, we follow \cite{style}, and define style as \textit{a distinctive manner which permits the grouping of works into related categories.} Nevertheless, instead of using hand-crafted features and predefined styles, we take data-driven probabilistic approach to determine stylistic similarity measure that we define in Sec.~\ref{subsec:blend}.

\section{Style Search Engine}
\label{sec:style_search_engine}

In this section, we present the pipeline of our multi-modal Style Search Engine. As an input, it takes two types of query information: an image of an interior, {\it e.g.} a picture of a dining room, and a textual query used to specify search criteria, {\it e.g.} {\it cozy and fluffy}. Then, an object detection algorithm is run on the uploaded picture to detect objects of classes of interest such as chairs, tables or sofas. Once the objects are detected, their regions of interest are extracted as picture patches and submitted to visual search method. 
Simultaneously, the engine retrieves the results for a textual query. With all visual and textual matches retrieved, our \textit{blending algorithm} ranks them depending on the similarity in the respective features spaces and serves the resulting list of stylistically and aesthetically similar objects. Fig.~\ref{pipe} shows a high-level overview of our Style Search Engine. Below, we describe each part of the engine in more details.

\subsection{Visual search}

Instead of using an entire image of the interior as a query, our search engine applies an object detection algorithm as a pre-processing step of. This way, not only can we retrieve the results with higher precision, as we search only within a limited space of same-class pictures, but we do not need to know the object category beforehand. This is in contrast to other visual search engines proposed in the literature~\cite{prod_des, pinterest}, where the object category is known at test time or inferred from textual tags provided by human labeling. 


As our object detection method, we use the state-of-the-art detection model YOLO 9000~\cite{yolo}. It is based on DarkNet-19 model~\cite{Darknet, yolo} with 19 convolutional layers and 5 max-pooling levels. YOLO 9000 is able to detect multiple furniture classes along with their bounding boxes. The bounding boxes are then used to generate Regions of Interest (ROIs) in the pictures and visual search is performed on the extracted ROIs.

In a set of initial experiments, we optimized the parameters of YOLO 9000 detection algorithm, mainly focusing on the detection confidence threshold. We set this threshold to $0.1$, although in case of overlapping bounding boxes returned by the model, we take the one with the highest confidence score.

Once the ROIs are extracted, we compute their representation using the outputs of pre-trained deep neural networks. More precisely, we use the outputs of fully connected layers of neural networks pre-trained on ImageNet dataset~\cite{russ}. We then normalize the extracted vectors of outputs, so that their $L_2$ norm is equal to $1$ and search for similar images within the dataset using this representation. To determine the neural network architecture providing the best performance, we conducted several experiments described in details in Sec.~\ref{subsec:visual_search}.

\subsection{Text query search}

To extend the functionality of our Style Search Engine, we implement a text query search that allows to further specify the search criteria. This part of our engine is particularly useful when trying to search for interior items that represent abstract concepts, such as {\it minimalism} or {\it Scandinavian style}. 

In order to perform such a search, we need to find the mapping from textual information to vector representation of the interior item. The resulting representation should live in a multi-dimensional space, where stylistically similar objects reside close to each other. We formulate this problem in the following manner. Let us first define $\mathbf{f} \in \mathbb{R}^n$ to be a vector representation of an item stored in the database and $(t_1,  t_2, \ldots, t_i ) = \mathbf{t} \in \mathfrak{T}$ be a variable length sequence that represents a textual query. We are interested in finding a mapping $m: \mathfrak{T} \rightarrow  \mathbb{R}^n$ from the space of queries to the vector space of interior items, such that $dist(m(\mathbf{t}), \mathbf{f})$ is small, when $\mathbf{f}$ are relevant to the query $\mathbf{t}$.
Having found such a mapping, we can perform search by returning $k$-nearest neighbors of transformed query in interior item space using cosine similarity as a distance measure. 

To obtain the above defined space embedding, we use a state-of-the-art Continuous Bag-of-Words (CBOW) model that belongs to word2vec model family~\cite{word2vec}. We use the descriptions of various household parts, such as living rooms or kitchens, to infer the contextual information about interior items. Such descriptions are available as part of the IKEA dataset which we describe in details in Sec.~\ref{sec:dataset}. It is worth noticing that our embedding is trained without relying on any linguistic knowledge since the only information that the model sees during training is whether given objects appeared in the same room. 

\begin{figure}[t!]
\includegraphics[width=0.5\textwidth]{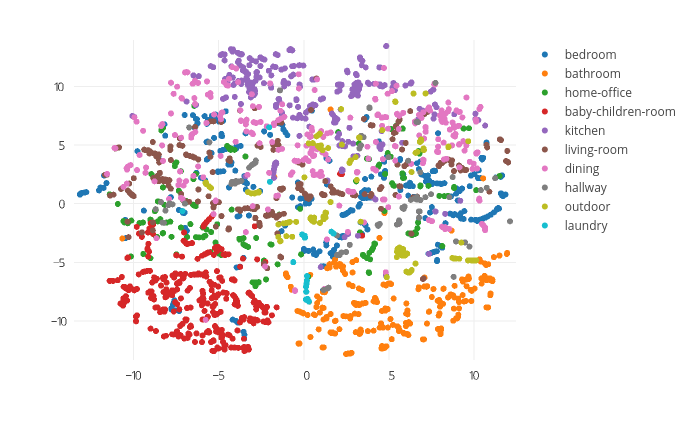} 
\vspace{-0.8cm}
\caption{t-SNE visualization of interior items' embedding. Distinctive classes of objects, {\it e.g.} those that appear in bathroom or baby room, are clustered around the same region of the space.}
\label{w2vec_tsne_embedding}
\end{figure}

In order to optimize hyper-parameters of CBOW for furniture embedding, we run a set of initial experiments on the validation dataset and use cluster analysis of the embedding results. We select the parameters that minimize intra-cluster distances at the same maximizing inter-cluster distance. Fig.~\ref{w2vec_tsne_embedding} shows the obtained feature embeddings using t-SNE dimensionality reduction algorithm \cite{tsne}. One can see that some classes of objects, {\it e.g.} those that appear in bathroom or baby room, are clustered around the same region of the space.

After obtaining the furniture embedding, we need a model to find an appropriate mapping $m: \mathfrak{T} \rightarrow  \mathbb{R}^n$ from query space to the space of furniture embeddings. 

To this end, we train a Long Short-Term Memory (LSTM) deep neural network architecture that has been successfully applied in several other natural language processing applications such as language modeling~\cite{jozefowicz2016}, machine translation~\cite{seq2seq} or on-line content popularity prediction~\cite{Stokowiec2017}. 

We formulate the question of finding  $m: \mathfrak{T} \rightarrow  \mathbb{R}^n$ as a regression problem. To be more explicit, let $\mathbf{t} = (t_1,  t_2, \ldots, t_i) \in \mathfrak{T}$ be a furniture description from IKEA Dataset and $\mathbf{f} \in \mathbb{R}^n$ denote its furniture embedding. We train our model to minimize the MSE between the predicted item embedding based on its description $\hat{\mathbf{f}} = LSTM(\mathbf{t})$ and the ground-truth furniture embedding $\mathbf{f}$.

Due to the fact, that vocabulary of IKEA Dataset products description is rather limited and may possibly not contain words from user-generated queries, we initialized the LSTM's query embedding layer  with word embeddings trained on dump of English Wikipedia with CBOW model. 
Additionally, to avoid overfitting, we froze the query embedding layer during training.

%

\section{Dataset}
\label{sec:dataset}

\begin{figure*}[t]
  \begin{center}
    \includegraphics[width=0.8\textwidth]{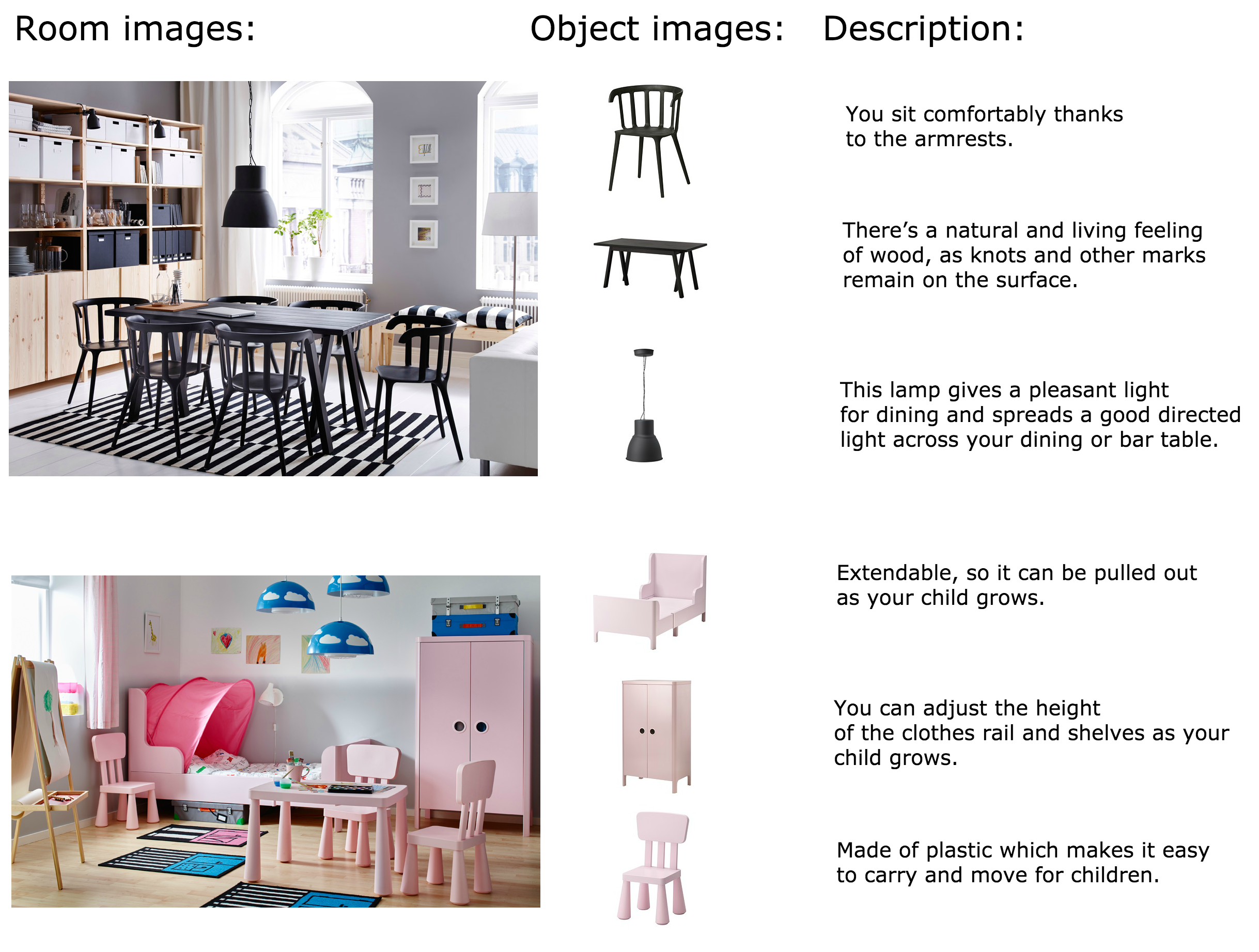} 
    \caption{Example entries from IKEA dataset contain room images, object images and their respective text descriptions.}
    \label{ikea_dataset_description}
  \end{center}
\end{figure*}

In order to evaluate our proposed Style Search Engine, we collected a dataset of interior items along with their textual description and the context in which they appear. Although several datasets for standard visual search methods exist, {\it e.g.} Oxford 5K \cite{philbin} or Paris 6K \cite{philbin2}, we could not use them in our work, as our multi-modal approach requires additional type of information to be evaluated. More precisely, our dataset that can be used in the context of multi-modal interior design search engine should fulfill the following conditions: 
 
\begin{itemize}
\item It should contain both images of individual objects as well as room scene images with those objects present.
\item It should have a ground truth defining which objects are present in a given room scene photo.
\item It should also have a textual description for each room scene image.
\end{itemize}

To our knowledge, no such dataset is publicly available. Hence, we collected our own dataset by recursively scrapping the website of one of the most popular interior design distributor - IKEA\footnote{\textit{https://ikea.com/}}. We were able to download 298 room photos with their description and 2193 individual product photos with their textual descriptions. A sample image of the room scene and interior item along with their description can be seen in Fig.~\ref{ikea_dataset_description}. We have also grouped together some of the most frequent object classes ({\it e.g.} chair, table, sofa) for more detailed analysis. In addition, we also divided room scene photos into 10 categories based on the room class (kitchen, living room, bedroom, children room, office). This kind of classification can be useful, {\it e.g.} for qualitative analysis of embedding results, as shown in Fig.~\ref{w2vec_tsne_embedding}. We plan to release our IKEA dataset to the public.

\section{Experiments}
\label{sec:experiments}

In this section, we present the results of the experiments conducted using our Style Search Engine to evaluate its performance with respect to the baseline methods. We first show how incorporating object detection algorithm and deep neural network architectures within our visual search engine improves the search accuracy. We then present our method for blending the results of multi-modal search and prove that using this approach we can increase the system performance by 11\%.

\subsection{Visual Search with Object Detection and Neural Networks}
\label{subsec:visual_search}

In this experiment, we analyze the results of our visual search when using various neural network architectures combined with YOLO 9000 object detection algorithm. The goal of this experiment is to select the right configuration of deep neural network used as the descriptor extractor for our interior design images, as well as to quantify the improvement obtained when adding a pre-processing step of object detection. To that end, we evaluate two neural network architectures that were successfully applied to object recognition task on ImageNet dataset: ResNet~\cite{HeZRS15} and VGG\cite{vgg}. We use VGG network with $3 \times 3$ convolutional filters in two configurations, with 16 and 19 weight layers. We analyze the outputs of the first (\textit{fc6}) and the second fully connected layer (\textit{fc7}) of the VGG network. For ResNet, we take the average pooling layer. In all experiments, we use normalized outputs of the networks pre-trained on ImageNet dataset and we compute the similarity measure with Euclidean distance. The networks were implemented using Keras~\cite{keras} with Theano backend for deep feature extraction.

%

{\bf Baseline:} As our baseline, we take the conventional Bag-of-Visual-Words search engine~\cite{vgoogle}. It is based on the SIFT feature extraction algorithm~\cite{sift}. We extract the descriptors and cluster them using $k$-means clustering~\cite{nister} into $k=1000$ {\it visual words}. We use SIFT implementation available in OpenCV for Python ~\cite{opencv} with contrast threshold set to $0.05$, edge threshold to $11$ and $L_2$ norm.

{\bf Evaluation metric:} To measure the performance of our system, we use Hit@$k$ metric~\cite{yt8m}. We define it in the following manner. Let $\mathcal{F}$ denote a set of all possible interior items available in the dataset. We define a room $\mathcal{R} \in \mathfrak{R}$ as a set that contains elements $f \in \mathcal{F}$. Hit@$k$ is therefore defined as the fraction of retrieved items that contain at least one of the ground truth objects in the top \textit{k} predictions. More formally, if $rank_{f,\mathcal{R}}$ is the rank of furniture $f$ in the room $\mathcal{R}$ (the highest scoring furniture having rank 1) and $G_{\mathcal{R}}$ is the set of ground-truth objects for $\mathcal{R}$, then Hit@$k$ is defined as: 
\begin{equation}
\frac{1}{|\mathfrak{R}|} \sum_{\mathcal{R} \in |\mathfrak{R}|} \vee_{f \in G_{\mathcal{R}}} \mathbb{I}(rank_{f, \mathcal{R}} \leq k),
\end{equation}
where $\vee$ is logical OR operator. 

\begin{figure}[t]
\includegraphics[height=140px]{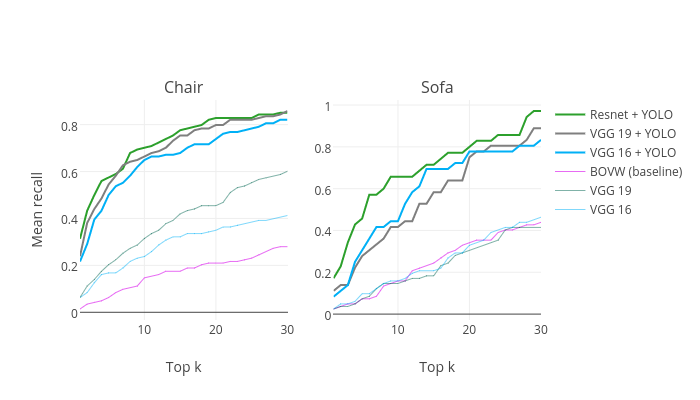} 
\vspace{-0.8cm}
\caption{Quantitative evaluation of various feature extraction methods combined with object detection algorithm YOLO 9000. We use recall as an evaluation metric that shows whether or not a single item present in the room picture was returned by the search engine. The recall is plotted  as a function of the number of returned items $k$.}
\label{chair_recall}
\end{figure}

{\bf Results:} Tab.~\ref{results_visualsearch} displays the results obtained for this experiment. Adding object detection algorithm as a pre-processing step significantly increases the number of correctly retrieved results across all evaluated configurations. We have illustrated the results for Hit@$6$ as we retrieved visually similar objects for six distinct object classes - chair, table, sofa, bed, wall clock and pottedplant. For Hit@$6$ the performance gains reach up to 175\% (in the case of ResNet) and 238\% (for VGG-19 with fc7). 
Feature extraction with ResNet and object detection pre-processing yields the highest Hit@$k$ score, retrieving correct results for almost half of all queries. To further analyze the performances of the proposed methods, in Fig.~\ref{chair_recall} we also plot recall curves for two sample object classes. Again, ResNet combined with object detection step remains the best performing configuration. One can also notice that all methods based on deep network architectures significantly outperform baseline BoVW method.

\begin{table}[t]
  \setlength{\tabcolsep}{6pt}
  \def\arraystretch{1.4}
  \centering
  \caption{Results for Content Based Image Retrieval experiment for different models and all object classes. Configuration of ResNet neural network with YOLO 9000 object detection as a pre-processing step significantly outperforms both the baseline BoVW model and other deep neural network architectures.}
  \label{results_visualsearch}
  \begin{tabular}{c|c|c|c}
  \toprule
  \multirow{2}{*}{\bf Model} 	& \multirow{2}{*}{\bf Layer}	& \multicolumn{2}{c}{\bf Hit@6 }	 \\ 
  \cline{3-4}
  & & {\it whole image} & {\it with object detection} \\
  \midrule
  BoVW 	& N/A 				   	& 0.066 & 0.26  \\ \midrule
  \multirow{2}{*}{VGG-16} & fc6 & 0.126 & 0.392 \\
  & fc7 & 0.153  & 0.314 \\ 
  \midrule
  \multirow{2}{*}{VGG-19} & fc6 & 0.141 & 0.43 \\
  & fc7 & 0.136  & 0.445 \\
  \midrule
  ResNet & avg pool & 0.167 & {\bf 0.458}\\  
  \bottomrule
  \end{tabular}
\end{table}

\subsection{Results blending}
\label{subsec:blend}

\begin{table*}[t]
\setlength{\tabcolsep}{6pt}
\def\arraystretch{0.95}
\centering
\caption{Mean similarity results averaged for all room pictures in IKEA dataset and sample text queries.}
\label{results_similarity}
\begin{tabular}{@{}lcccc@{}}
\toprule
\multirow{2}{*}{Text query} 			& \multirow{2}{*}{Visual search}	    & \multirow{2}{*}{Text search} & \multirow{2}{*}{Simple blending}  	&   Feature similarity \\
& & & & blending \\
 \midrule
-					& 	0.2295 		 & -	& -					& - \\ 
object class name	& 	-  			 &  -	& 0.2486 			& 0.2374	\\ 
 \textit{decorative}  &  - 			& 0.1358	& 0.2316 			& 0.2517	\\ 
 \textit{black}		& - 		 	& 0.1538	& 0.2493  			& 0.2244	\\ 
\textit{white}		& - 		 	& 0.2036	& 0.2958  			& 0.2793	\\ 
 \textit{smooth} 	& - 		 	& 0.3520	& 0.2415  			& 0.3052	\\ 
\textit{cosy}		& - 		 	& 0.2419	& 0.2126  			& 0.2334 	\\ 
\textit{fabric}		& - 		 	& 0.0371	& 0.1269  			& 0.1344 	\\ 
\textit{colourful}	& -  		 	& 0.4461	& 0.3032  			& 0.3215	\\ \midrule
Average 			& 0.2295        & 0.2243	& 0.2387  			& \textbf{0.2484}   \\
\bottomrule
\end{tabular}
\end{table*}


In order to use the full potential of our multi-modal interior design search engine, we introduce a blending method to combine the retrieval results of visual and textual search engines and present them to the user. To that end, we use {\it feature similarity blending} approach. More precisely, the search engine returns an initial set of results for each modality, extracts visual features (normalized outputs of pre-trained deep neural network) and then re-ranks them using the distance from the query to the item in visual features' space for each modality independently (visual search results do not need to be re-ranked). A set of closest items is returned as a final result. 

{\bf Simple blending:} As an alternative method for blending the results, we blend $k$ best results from each modality and return them as a final result. 



{\bf Evaluation metric:} As mentioned in Sec.~\ref{subsec:stylistic_similarity}, defining a similarity metric that allows to quantify the stylistic similarity between interior design objects is a challenging task and an active area of research. In this work, we propose the following similarity measure that is inspired by~\cite{style} and based on a probabilistic data-driven approach. Similarly to Hit@$k$ metric, let us first define $\mathcal{F}$ as a set of all possible interior items available in our dataset and a room $\mathcal{R} \in \mathfrak{R}$ as a set containing elements $f \in \mathcal{F}$. Our proposed similarity metric between two items $f_1$, $f_2 \in \mathcal{F}$ that determines if they fit well together can be computed as:

\begin{equation}
	C(f_1, f_2) =| \{ \mathcal{R} : f_1 \in \mathcal{R} \wedge  f_2\in \mathcal{R} \}|.
\end{equation}
We defined the style similarity as:  
\begin{equation}
	s(f_1, f_2) = \frac{C(f_1, f_2)}{ \max_{ f_i, f_j \in \mathcal{F} } \; C(f_i, f_j)}.
\end{equation}
In fact, it as the fraction of the number of rooms, in which both  $f_1$ and $f_2$ appear and total number of rooms in which any of those items co-occur. This metric can be interpreted as empirical probability for two objects $f_1$ and $f_2$ to appear in the same room.


{\bf Results:} Tab.~\ref{results_similarity} shows the results of the blending methods in terms of mean value of our similarity metric. \textit{Text query = object class name} means that detected object class, {\it i.e.} the one with the highest detection confidence, was used as a text query.

Vanilla visual search without text query achieves an average value of $0.2295$ where similarity is calculated over visually similar results to the query object, all belonging to the same object class. For text search average similarity was slightly lower - $0.2243$.

When analyzing the results of the evaluated blending approaches, both of them have a score that is higher than the ones obtained for vanilla visual and text search. Our proposed blending method outperforms both the visual search and simple blending, yielding an improvement of 11\% and 4\% respectively. It is worth noticing that simply adding a name of detected object class as a text query  improves the search results already. Providing additional information such as color or style ({\it e.g.} \textit{white} or \textit{decorative}) yields further performance improvement.

%

\section{Web Application}
\label{sec:application}

\begin{figure*}[t]
\begin{center}
\includegraphics[width=0.8\textwidth]{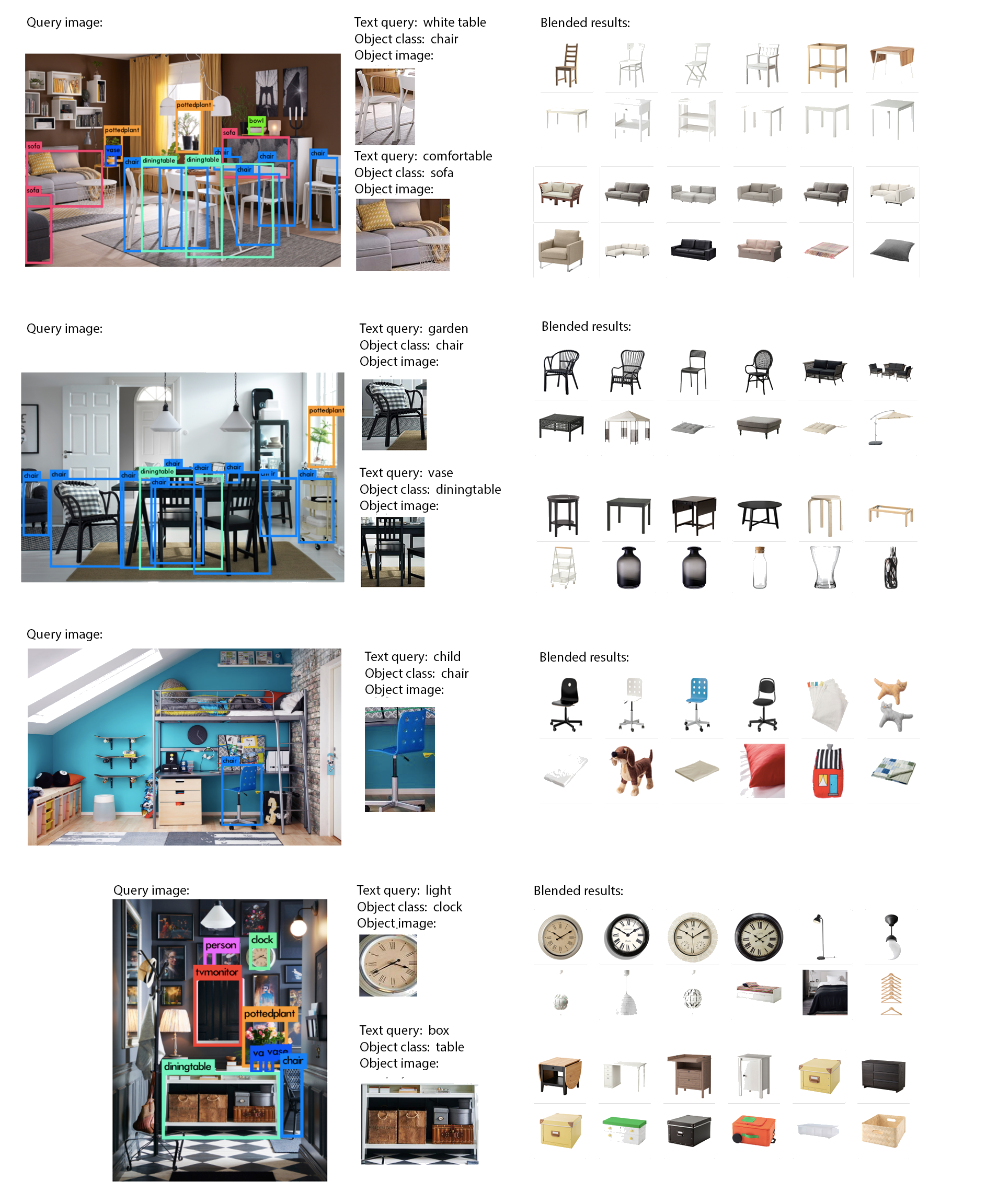} 
\caption{Screenshots of the Web application of our Style Search Engine. Sample results retrieved for room images from the IKEA dataset and combined with custom text queries.}
\label{blended_results}
\end{center}
\end{figure*}

To enable dissemination of our work, we implemented a Web-based application of our Style Search Engine. The application allows the user either to choose the query image from a pre-defined set of room images or to upload his/her own image. The application was implemented using Python Flask\footnote{http://flask.pocoo.org/} - a lightweight server library. It is currently available for restricted use only\footnote{http://style-search.us-west-2.elasticbeanstalk.com} and we plan to open it to the public, once it passes the initial tests with trial users. Fig.~\ref{blended_results} shows a set of screenshots from the working Web application with Style Search Engine.

\section{Conclusions}
\label{sec:conclusions}
In this paper, we proposed a multi-modal search engine for interior design applications dubbed Style Search Engine. By combining textual and visual information, it can successfully and with high recall retrieve stylistically similar images from a dataset of interior items. Thanks to the object detection pre-processing step, the results of our visual search component improved by over 200\%. Using feature similarity blending approach to combine the results of visual and textual search engines, we increased the overall similarity score of the retrieved results by 11\%. We also implemented working prototype of a Web application that uses our Style Search Engine.

In our future research, we plan to explore various approaches towards common latent space mapping that could allow to map both textual and visual queries to a common space and perform similarity search there.




\bibliographystyle{ieeetr}
\bibliography{bibliography.bib}



\end{document}